\documentclass[10pt,twocolumn,letterpaper]{article}

\usepackage{wacv}
\usepackage{times}
\usepackage{epsfig}
\usepackage{graphicx}
\usepackage{amsmath}
\usepackage{amssymb}
\usepackage{caption}
\usepackage{algorithm}
\usepackage{algorithmic}
\usepackage{xcolor}

\usepackage[bookmarks=false, colorlinks=true]{hyperref}

\wacvfinalcopy 

\def\wacvPaperID{117} 

\ifwacvfinal\fi
\begin{document}

\title{Learning Video-Story Composition via Recurrent Neural Network}

\author{Guangyu Zhong$^{1, 2}$
	\hspace{0.08in} Yi-Hsuan Tsai$^{3}$
	\hspace{0.08in} Sifei Liu$^4$
	\hspace{0.08in} Zhixun Su$^1$
	\hspace{0.08in} Ming-Hsuan Yang$^2$\\
	\hspace{0.15in} $^1$School of Mathematical Sciences, Dalian University of Technology \\
	\hspace{0.15in} $^2$Electrical Engineering and Computer Science, University of California, Merced \\
	\hspace{0.15in} $^3$NEC Labs America \hspace{0.15in} $^4$NVIDIA Research
}

\maketitle
\ifwacvfinal\thispagestyle{empty}\fi

	\begin{abstract}
	In this paper, we propose a learning-based method to compose a video-story from a group of video clips that describe an activity or experience.
	We learn the coherence between video clips from real videos via the Recurrent Neural Network (RNN) that jointly incorporates the spatial-temporal semantics and motion dynamics to generate smooth and relevant compositions.
	We further rearrange the results generated by the RNN to make the overall video-story compatible with the storyline structure via a submodular ranking optimization process.
	Experimental results on the video-story dataset show that the proposed algorithm outperforms the state-of-the-art approach.
\end{abstract}

\section{Introduction}
\label{sec:intro}
\begin{figure}\footnotesize
	\centering
	\includegraphics[width=1\linewidth]{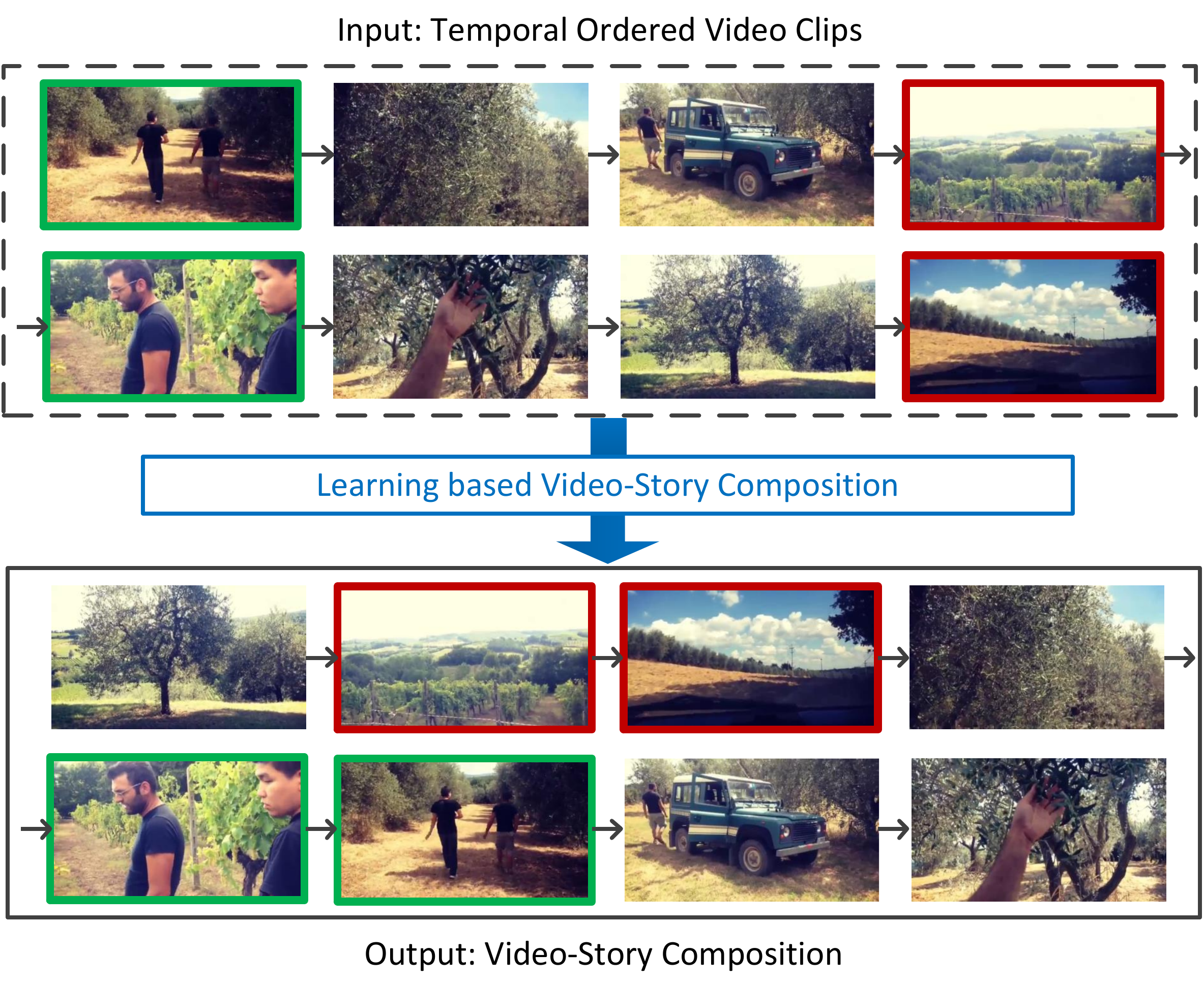}\\
	\caption{Learning video-story compositions. Given a set of temporally ordered video clips, our method learns to reorder and compose the clips into a coherent video-story that matches the storyline structures. 
		For example, given the video clips taken by a person while walking in the garden, the proposed method reorders the temporally ordered one and generates the coherent video-story, where the video clips with similar scenes (red rectangles) or contents (green rectangles) are connected together.}
	\label{fig:intro}    
\end{figure}
Nowadays people are able to capture and store more and more personal experiences and memories in videos with the decreasing cost of cameras and storages. 
To organize these captured videos, they are usually edited and processed to be a concise format at a later time.
Since the manual post-processing is time-consuming and labor-intensive, automatic algorithms are developed to process these unorganized videos, e.g., generation of a ``short story'' from a collection of videos \cite{choi_CVPR_2016_video}.
In this work, we adopt this problem setting and aim to composite a smooth and meaningful video-story from video clips.

Specifically, we describe our task as: given a set of clips taken by a person during an activity or experience, we find out an order of the clips which composes a story containing smooth transitions in terms of semantics, motions, and activity dynamics that match the storyline structures \cite{choi_CVPR_2016_video} (see Figure~\ref{fig:intro}). 
Note that, different from the video summarization task that aims to select keyframes out of a long video \cite{wolf_1996_key,zhang_1997_integrated,liu_2010_hierarchical,lee_2012_discovering}, the ``story composition'' problem described in this paper considers transitions between selected subshots and produces the consistent story in the temporal domain.

Recently, numerous methods address the temporal consistency problem by identifying temporal alignments \cite{basha_2012_photo,kim_2014_reconstructing}, storyline graph \cite{kim_2014_reconstructing} or learning temporal relations \cite{sigurdsson_ECCV_2016} from images, in order to make the story more meaningful.
%
%
However, most temporal alignment based methods suffer from two difficulties in practice.
First, the results may look incoherent when the story is extracted from multiple video clips taken at different times.
Second, the ambiguous scene transition of shots also affects the overall quality of the composition.


In order to solve the above challenge, hand-crafted features can be used to represent the relations between video clips. State-of-the-art method \cite{choi_CVPR_2016_video} uses the color based bidirectional similarity to describe the relations between video clips and the dense optical flow to generate dynamics scores for each clip. The video sequence order is then formulated and generated via a branch-and-bound algorithm \cite{horst_2013_global}.
However, the coherence between clips is built directly through feature matching, which is likely to fail in the cases when there are ambiguous appearances or interrupted motions. 
%
%

In contrast, we address this problem by modeling the coherence of adjacent clips 
through a learning-based recurrent network.
Our network learns how to select the next connected clip from the remaining set of video clips based on previous selections in the temporal domain. 
%
Specifically, we train the two-stream RNN, including a semantic RNN that uses the spatial-temporal features, and a motion RNN that exploits the motion dynamics in each video clip.
To train this network, a generated initial clip is fed into the streams, and two output probabilities are jointly fused as the coherence scores between video clips to predict the next clip.

We further rearrange the probabilities from the two-stream RNN by a submodular ranking process to align with the storyline structure, which consists of the exposition, rising, action, climax, and resolution. Generally, the storyline structure ensures that video-story contains rising dynamics and has an ending with more activity than its beginning to attract the viewers \cite{choi_CVPR_2016_video}.
Finally, we compose the video-story by solving this submodular ranking optimization.

We demonstrate the effectiveness of the proposed learning based video-story composition algorithm on the benchmark dataset \cite{choi_CVPR_2016_video}.
We conduct a user study via Amazon Mechanical Turk to evaluate the overall video-story quality and quantitatively verify the composition quality based on pairwise annotations.
Overall, our experimental results show that the proposed learning based algorithm performs favorably against the state-of-the-art methods in terms of visually quality and accuracy.

The main contributions of this work are summarized as follows.
First, we propose a novel learning-based framework via the two-stream RNN for video-story composition.
Second, we show that the proposed model explicitly learns better representations to model the coherence between video clips.
Third, we develop a submodular ranking algorithm to improve the video-story composition results that better match the storyline structure. 

\section{Related Work} 
{\flushleft {\bf Video Summarization.}}
As introduced in Section~\ref{sec:intro}, although having different goals, the technical aspects of video summarization are quit similar and can be sufficiently utilized by video composition.
Many video summarization approaches have been proposed via different image-based feature representations and optimization methods, either through low-level feature such as optical flow \cite{wolf_1996_key} and image differences \cite{zhang_1997_integrated}, or high-level representations, including object trackers \cite{liu_2010_hierarchical} and importance scores \cite{lee_2012_discovering}. 
On the other hand, subshot-based methods represent summarizations via spatio-temporal features \cite{liu2008human}.
Numerous supervised approaches select the subshots to represent the videos based on submodular function \cite{gygli_CVPR_2015} and exemplas \cite{zhang_2016_summary}. 
All these methods require ground truths for training. 

However, the labeling of the ground truth for either video summarization or video caption is too subjective and difficult as a consistent limitation to the above methods.
In contrast, our model is learned in an unsupervised manner, making the framework more flexible to utilize large amount of data to improve the performance.
{\flushleft {\bf Story Composition.}}
The story composition methods typically focus on identifying the temporal alignment of the image sets (photo albums).
Basha et al. \cite{basha_2012_photo} use static and dynamic features to find the temporal order of the image sequence. 
Kim et al. \cite{kim_2014_reconstructing} learn the pairwise transition to construct the storyline graphs.
Recently,  an unsupervised method proposed by Sigurdsson et al. \cite{sigurdsson_ECCV_2016} use a skipping Recurrent Neural Network to learn long-term correlations.

In contrast, our approach focuses on compositing video clips rather than images. The video clips contain significant dynamics and ambiguity in terms of semantics and motions, thereby resulting in more challenging scenarios.
In this work, we aim to rank all the video clips and compose a coherent story rather than selecting a subset of images or videos.
We note that our approach is closest related to the plot analysis based method \cite{choi_CVPR_2016_video}. 
Instead of solving this problem via hand-crafted features, we learn the coherence between clips from real videos.
{\flushleft {\bf Learning Temporal Representations.}}
Temporal representations have been used in many tasks in language analysis \cite{mikolov2010recurrent,sutskever2011generating} and computer vision \cite{Tran_ICCV_2015,sigurdsson_ECCV_2016}. 
Recurrent neural networks are used in language modeling \cite{mikolov2010recurrent} and text generation tasks \cite{sutskever2011generating} to analyze the temporal information across time steps and generate future contents.
Sigurdsson et al. \cite{sigurdsson_ECCV_2016} extend this idea by modeling long-term memories to represent each story topic. 
On the other hand, spatial-temporal information such as C3D features \cite{Tran_ICCV_2015} are used in  video analysis tasks \cite{yao_CVPR_2016,pan_CVPR_2015}. These features describe the temporal representation for activities through a set of video frames.
In this paper, we utilize these representations but focus on analyzing relations between video clips that contain various topics (e.g., different scenes or objects).

%
\section{Learning Video-Story Composition}
\begin{figure}\footnotesize
	\centering
	\includegraphics[width=0.9\linewidth]{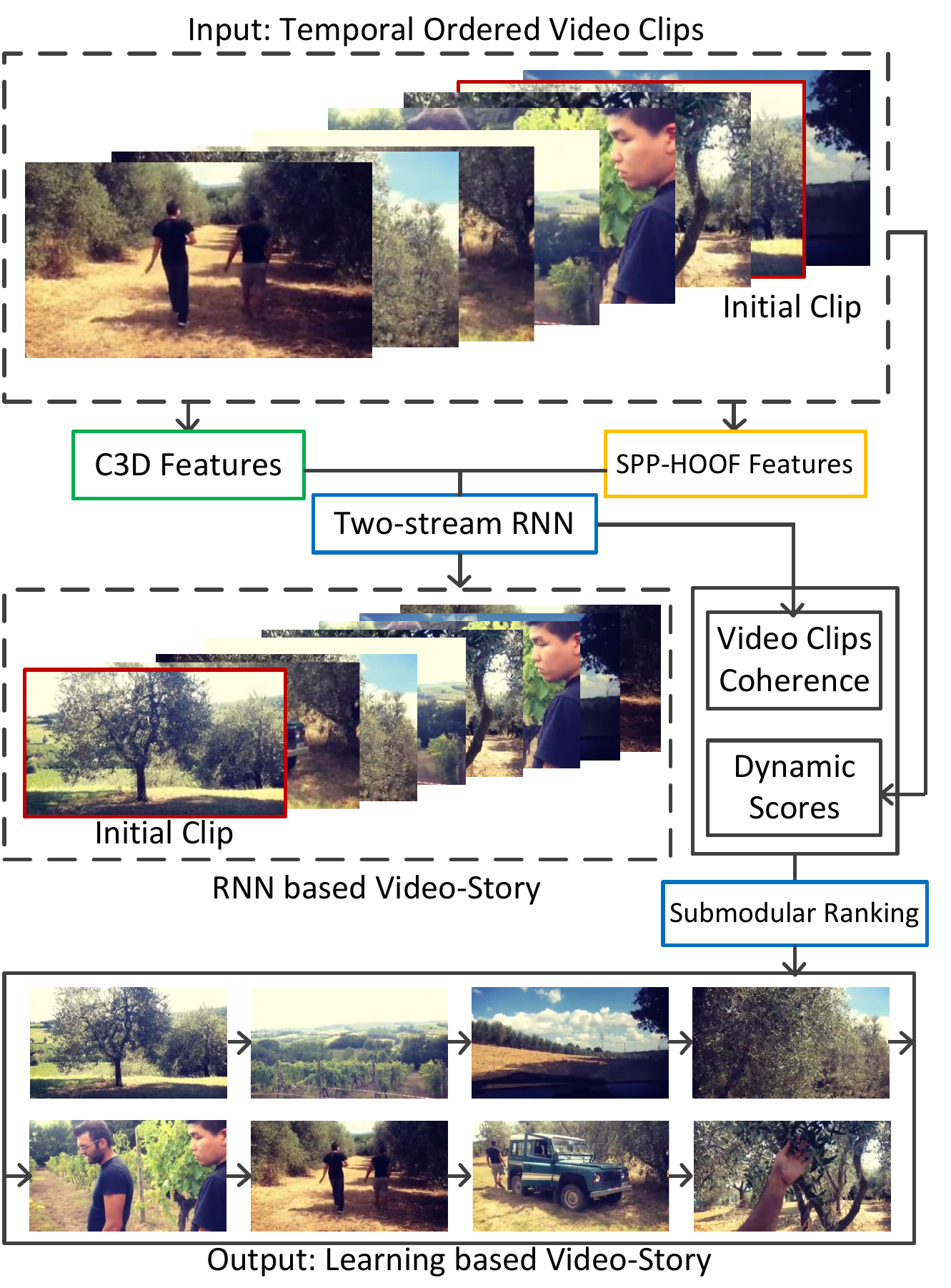}\\
	\caption{Overview of the proposed algorithm. We first feed the initial clip (in red rectangles) into the two-stream RNN. The output probabilities are then used as coherence scores between video clips to predict the next clip and generate the video-story.
		To further refine the results and match the storyline structure, we rearrange the composition order by solving a submodular ranking optimization via the learned coherence and activity dynamics of video clips.
	}
	\label{fig:overview}    
	\vspace{-2mm}
\end{figure}
\subsection{Overview}

Given a set of individual video clips, our goal is to compose the clips as a video-story which meets two criteria: (1) the semantic and motion transitions of the connected clips are coherent and smooth; (2) the composed video follows the storyline structure.
%
To achieve this, we first learn the coherence between video clips by training RNNs in an unsupervised manner.
%
We train a two-steam RNN with clip representations of the C3D features \cite{Tran_ICCV_2015} and optical flow. 
Then the probabilities generated in each RNN are jointly fused and learned to output the coherence score between clips.
To make the video-story match the storyline structure, we further model the video composition task as a ranking problem via a submodular optimization function guided by the learned coherence and activity dynamics of video clips.
Figure \ref{fig:overview} shows the main steps of the proposed algorithm.
\subsection{Learning Video Coherence via RNNs}
\begin{figure*}\footnotesize
	\centering
	\includegraphics[width=1\linewidth]{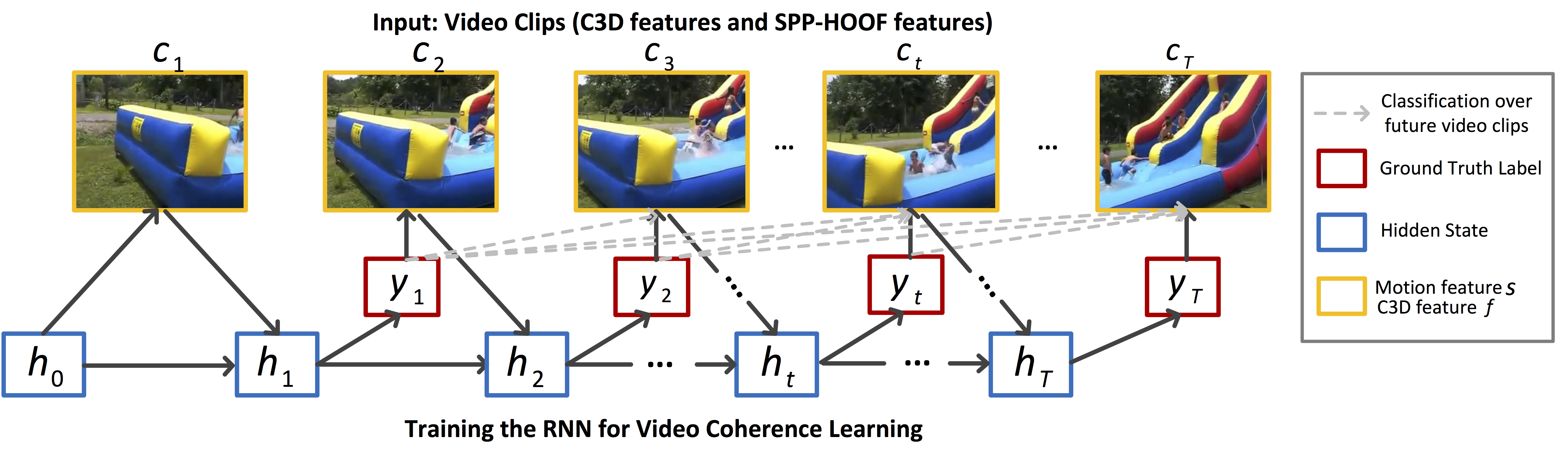}\\
	\caption{Illustration of the proposed RNN for learning the coherence between clips, where the hidden layer preserves the information from previous states.
		In the RNN model, we fix the length of training data as $T$ and treat the problem as a classification task with the soft-max function to predict the next frame from the remaining clips.
		Note that, we use the same architecture for our two-stream framework, where the inputs are C3D and SPP-HOOF features, respectively.
	}
	\label{fig:rnn}    
	\vspace{-2mm}
\end{figure*}
\label{section:coherence}
{\flushleft {\bf Recurrent Neural Networks.}} 
The RNN \cite{elman1990finding} can be used to process sequential data of a input video, which meets the need of the formulated task that aims to sequentially predict the next clip given previous contents.
Based on a series of $T$ items $c_{1:T} = \{c_1, ..., c_T\}$, where each $c$ represents a clip, the network is trained to predict the next clip by maximizing the log-likelihood:
\begin{equation}
\theta^* = \underset{\theta}{\text{argmax}}{\sum_{t}^{T-1}{\log P(c_{t+1}|c_{1:t}; \theta)}},
\end{equation}
%
where $\theta$ indicates all the parameters in the model. We use the back propagation through time method \cite{williams_1995gradient,werbos_1988generalization} to optimize the model.

The RNN model consists of the input, recurrent, and hidden layers. At the $t$-th time step, the output feature $y_t$ is computed as follows:
\begin{equation}
\begin{aligned}
h_t &= \sigma_h(W_Ic_t + W_Hh_{t-1}),\\
y_t &= \sigma_y(W_Oh_{t}).
\label{eq:rnn}
\end{aligned}
\end{equation}
The input $c_t$ is used to update the hidden recurrent layer $h_t$ with the weights $W_I$, and the hidden layer updates itself using the weights $W_H$. Then the output $y_t$ is generated via weights $W_O$ and non-linear activation functions $\sigma_h$ and $\sigma_y$. 

{\flushleft {\bf Loss Function.}} 
Considering that different topics may appear in video clips, it is not trivial to obtain video composition ground truths, we formulate an unsupervised learning task using the temporal order in real videos.
We first split the entire video into several video clips, and each clip contains fixed length of frames. To avoid the vanishing gradient problem \cite{bengio_1994learning} introduced by training the long-term data, we randomly select a continuous subset of clips for training rather than directly using the entire set.
Note that the length of input video clips $c_{1:T}$ is fixed.
%

Based on the previous chosen $t$ clips $c_{1:t}$, we wish to choose the next clip $c_{t+1}$ from the remaining clips $\mathcal{C}_t = c_{t+1:T}$ using maximum likelihood. We define the probability of an unselected clip $c_{\tau} \in \mathcal{C}_t$ being the next selected clip $c_{t+1}$ using the soft-max function over the inner product of the network output $y_t$ and the feature vector of $c_{\tau}$:
%
\begin{equation}
P(c_{\tau} = c_{t+1}|c_{1:t};\theta) = \frac{\exp(y_{t}^{\top}c_{\tau})}{\sum_{c\in \mathcal{C}_t}\exp(y_{t}^{\top}c)}.
\end{equation}
Different from the standard language model \cite{mikolov2010recurrent} that directly generates representations of the next clip, this formulation selects the best matching item among the remaining ones.
%
We illustrate the architecture of our model in Figure \ref{fig:rnn}.
{\flushleft {\bf Representation for Video Clips.}}
To describe the spatial-temporal information within each clip, the C3D features \cite{ji_PAMI_2013, Tran_ICCV_2015} have been effectively used to represent the clip.
Specifically, we utilize a C3D model pre-trained on the Sports-1M video dataset \cite{karpathy_CVPR_2014}, and extract features form the fc7 fully-connected layer as the representation $f$ for an input clip $c$.

To describe the motion dynamics in each video clip, we first extract the dense optical flow \cite{liu_2011_sift} from each frame, which is shown to provide effective representations 
for action recognition  \cite{donahue_CVPR_2015}. 
Then we compute the histogram of dense optical flow (HOOF) \cite{chaudhry_CVPR_2009} to generate a feature vector.
Since the motions in each frame may vary significantly at different locations (e.g., the background scenes usually contain fewer actions compared with the foreground objects), we further adopt the spatial pyramid pooling (SPP) \cite{he_ECCV_2014} on the optical flow to generate an SPP-HOOF feature for each frame.
Given a video clip $c$ with $l$ frames and the pyramid level $\{M \times M\}$, 
the SPP-HOOF feature $s^k$ in the $k$-th frame is defined as: $s^k = [h^{k_1}, ..., h^{k_{(M \times M)}}, h^{k}]$. 
We then normalize our SPP-HOOF motion features in clip $c$ as
$s = \frac{1}{l}{{\sum_{k=1}^{l}{s^k}}}.$

{\flushleft {\bf Learning Coherence between Clips.}}
Motivated by \cite{simonyan2014two}, we train the two-steam RNN that accounts 
for semantics and motions using the above-mentioned representations for clips (i.e., C3D features $f$ and SPP-HOOF features $s$).
%
To train this network, an initial clip is required.
%
Since a good video story that matches the storyline structure usually starts with a clip that contains fewer motions \cite{choi_CVPR_2016_video}, we compute a dynamics score $\phi$ as the average magnitude of the optical flow in each clip and select the one with the smallest score as the initial clip.

Given $N$ video clips, at each training step $t$, we fuse the output probabilities from the two streams to describe the coherence between previously ordered clips ${c}_{1:t}$ and the remaining ones ${\mathcal{C}}_t$. 
Thus the corresponding coherence vector $d({c}_t, c)$ is defined as:
\begin{equation}
d({c}_t, c) = \{ \lambda P(f|{f}_{1:t};\theta_f) + (1-\lambda) P(s|{s}_{1:t};\theta_s), c \in \mathcal{{C}}_t\},
\label{eq:distance}
\end{equation}
where $\theta_f$ and $\theta_s$ are the parameters in the semantic and motion streams respectively, and $\lambda$ is set to 0.5 for averaging the probabilities.
We consider this process as our baseline method, in which the next video clip is the one with the highest coherence score.

\begin{figure}\footnotesize
	\centering
	\includegraphics[width=0.95\linewidth]{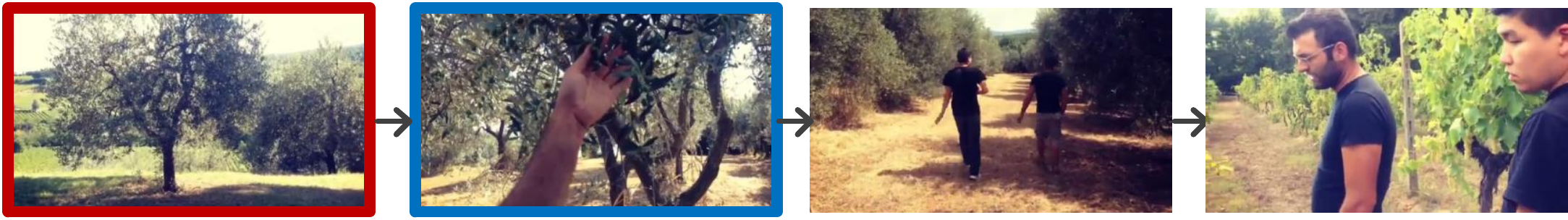}\\
	(a) Video composition by matching the similarities of C3D features.
	\includegraphics[width=0.95\linewidth]{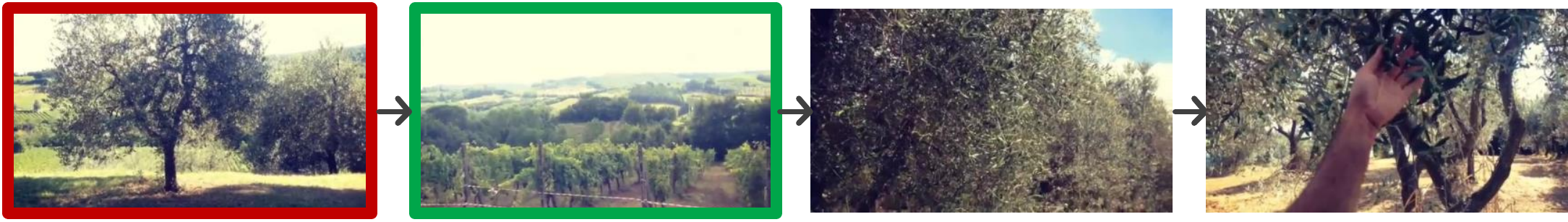}\\
	(b) Video composition by our semantic RNN with one unrelated clip.
	\includegraphics[width=0.95\linewidth]{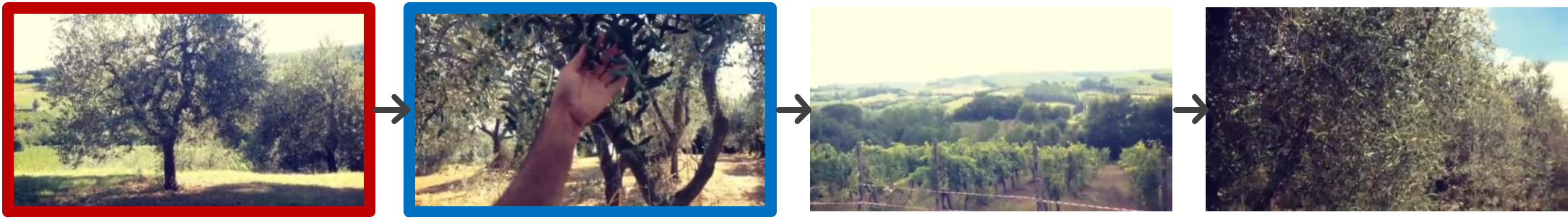}\\
	(c) Video composition by our semantic RNN.
	\caption{Effects of the coherence learned by RNN. Given the initial video clip (marked with red rectangle), we show the selected video clips in (a) using direct feature matching and (b) using our semantic RNN. 
		It shows our RNN emphasizes holistic rather than local coherence in the temporal space, and results in a more consistent composition. 
	}
	\label{fig:match}    
	\vspace{-2mm}
\end{figure}

To validate the effectiveness of our learned coherence in terms of the semantics, we analyze the results generated by our semantic RNN in Figure \ref{fig:match}.
Figure \ref{fig:match} (c) shows that with our semantic RNN, the scene of forests (the red rectangle) is followed by similar scenes (e.g., the green rectangle in (c)) rather than unrelated scenes or activities (e.g, the blue rectangle in (a)), which provides smoother transitions. 
Furthermore, our transitions are robust. Even with one unrelated clip inserted (the blue rectangle in (b)), the story of following clips are not heavily affected by this clip due to the accumulated information learned by our semantic RNN, and thus the consistency of the whole story is kept. 
%

\begin{figure}\footnotesize
	\centering
	\begin{tabular}{c@{\hspace{1mm}}c@{\hspace{1mm}}c@{\hspace{1mm}}c@{\hspace{0.1mm}}c@{\hspace{0.1mm}}c@{\hspace{0.1mm}}c@{\hspace{0.1mm}}c@{\hspace{0.1mm}}c}        
		\includegraphics[width=0.32\linewidth]{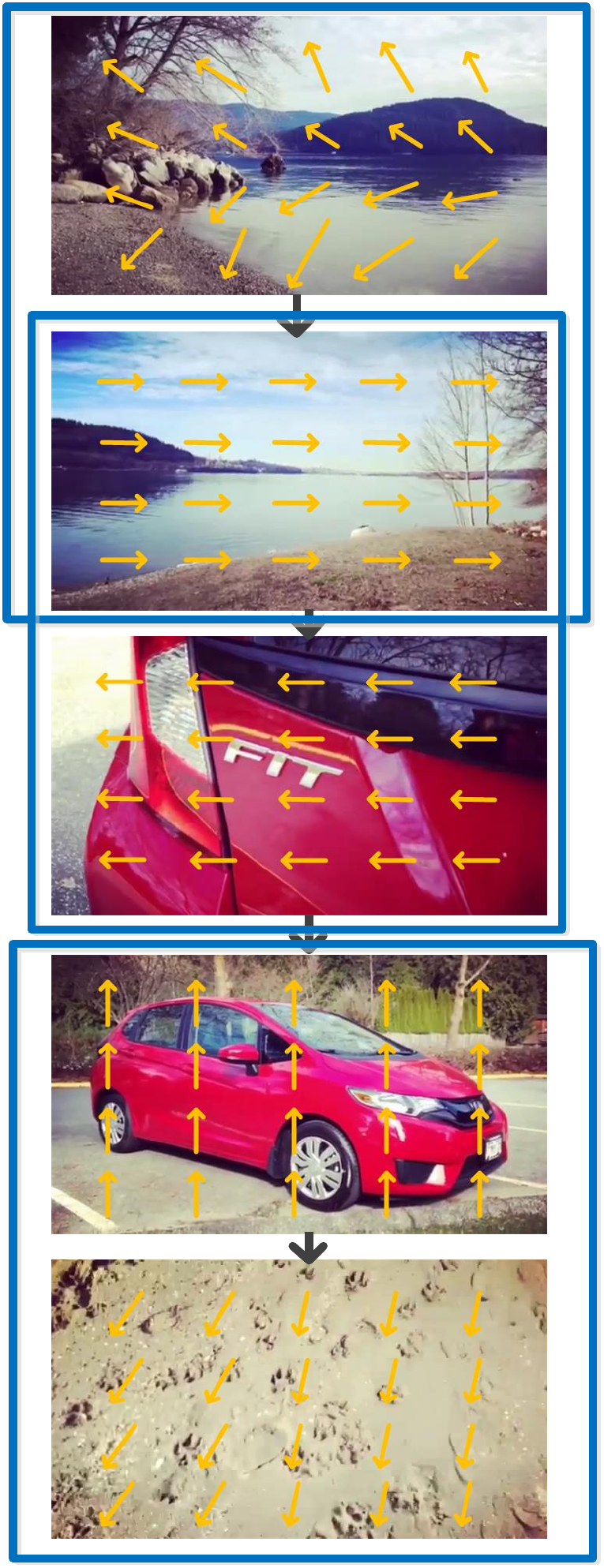}&
		\includegraphics[width=0.275\linewidth]{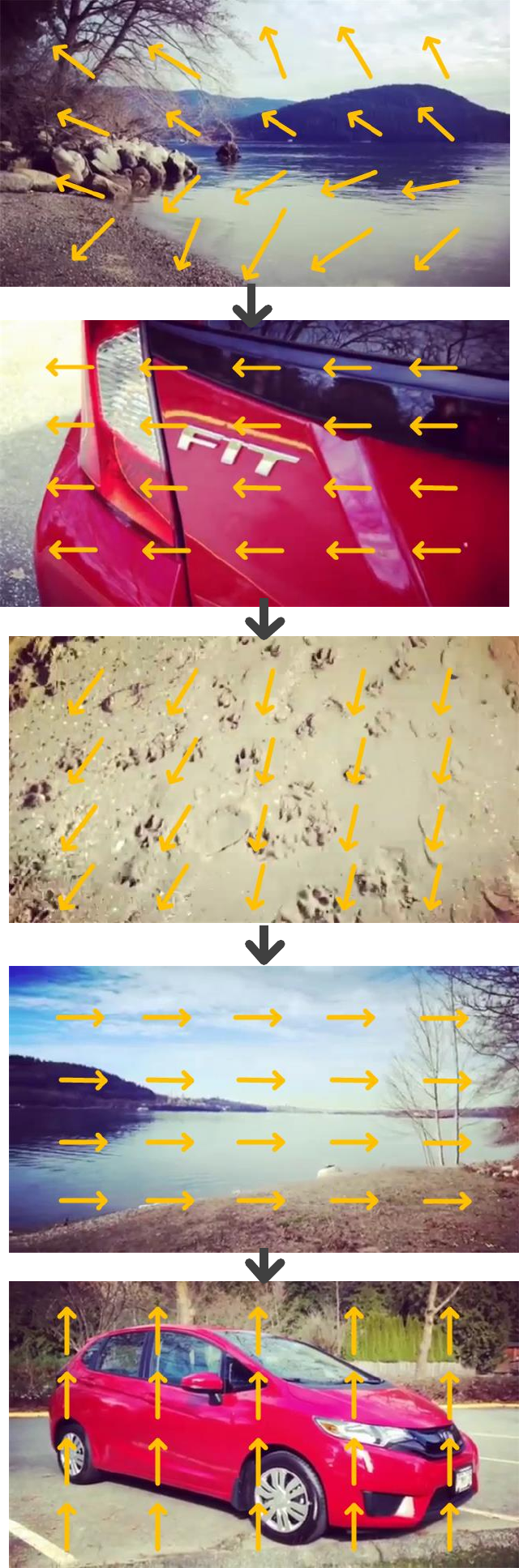}&
		\includegraphics[width=0.323\linewidth]{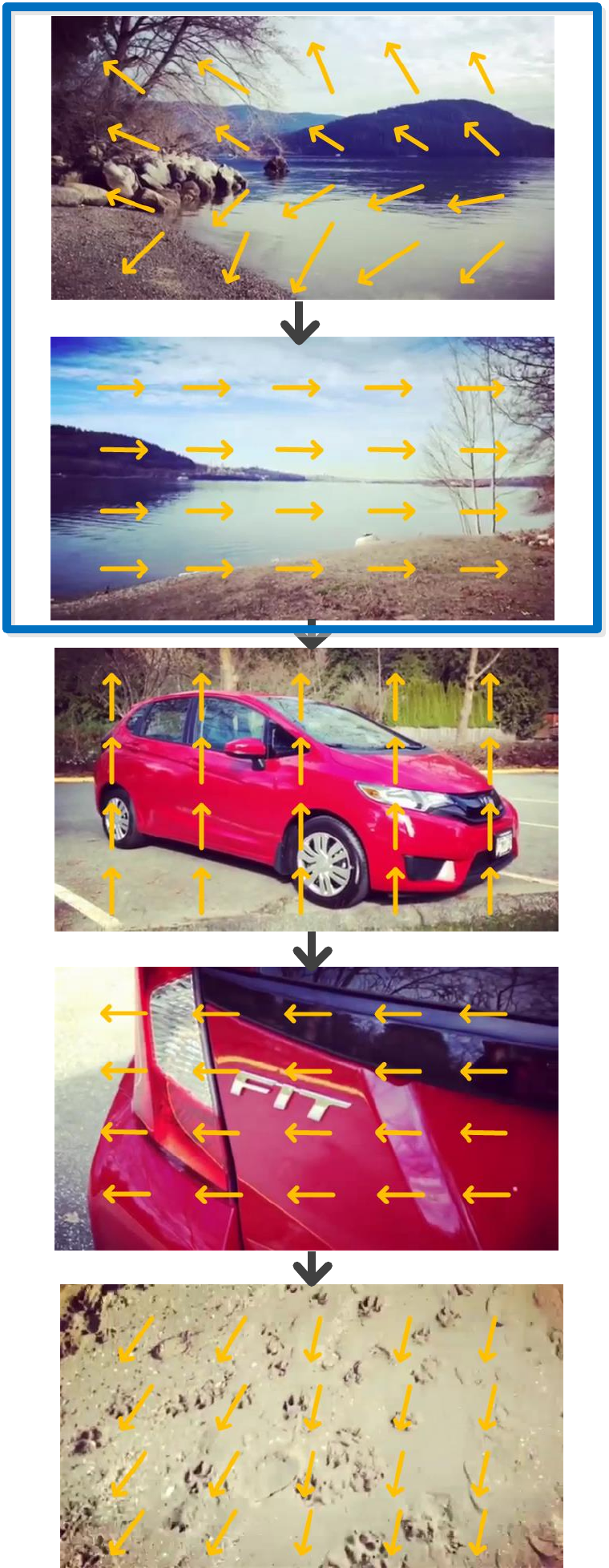}
		\\
		(a) Semantic RNN & (b) Motion RNN  & (c) Two-stream RNN
	\end{tabular}
	\caption{ Video composition by different RNNs. Given the same initializations, (a) and (b) are generated via the semantic RNN and motion RNN, respectively. (c) is generated by our baseline. 
		The contents in (a) are consistent while the motion transitions change a lot, (e.g., the major motions of clips in the blue rectangles change about 180 degrees). The motion RNN provides better motion consistency in (b) while the adjacent contents are less related.
		Our two-stream RNN in (c) contains more consistency in terms of both semantics and motions. 
		The major motion directions of each clip are shown by orange arrows.
	}
	\label{fig:rnn-result}
	\vspace{-2mm}
\end{figure}

%
In addition, without considering the motion stream, the results of the single semantic RNN may contain significant motion change across the adjacent clips that could potentially cause motion sickness. For example, the major motions of the composition by the semantic RNN in Figure \ref{fig:rnn-result} (a) flip (i.e., the motion direction of the camera changes almost 180 degrees) three times (marked in the blue rectangles) while the ones in Figure \ref{fig:rnn-result} (c) only flip once via merging the motion consistency. The quantitative results in Figure \ref{fig:curve} further validate the effectiveness of our coherency.


\subsection{Submodular Ranking}
\label{sec:submodular}
To ensure the video-story composition meets the storyline structure, we formulate a submodular optimization problem to select and rearrange video clips from the ordered set generated by the two-stream RNN. 
We first construct a graph where video clips are considered as nodes. 
We design a submodular objective function using the coherence and activity dynamics of video clips to describe the ideal video-story. The video-story result is then extracted by solving this proposed submodular function.
{\flushleft {\bf Graph Construction.}} 
Given a set of ordered video clips $\tilde{c}_{1:N}$ generated by our the proposed two-stream RNN, we construct a fully connected graph $G=(\mathcal{V}, \mathcal{E})$.
Each element $v \in \mathcal{V}$ is a video clip from $\tilde{c}_{1:N}$ and the edge $e\in\mathcal{E}$ represents the pairwise relation between two clips.
We aim to select all the nodes from $\mathcal{V}$ to $\mathcal{A}$. Then the selection ranks are considered as the composition order of video clips $\mathcal{A}$. 

{\flushleft {\bf Submodular Function.}}
We aim to select the video clip that meets two criteria: (1) sharing high coherence with other clips; (2) providing rising activity dynamics.  
The objective function is formulated with two terms, i.e., the facility location (FL) term to describe the coherence between candidate clips, and an activity dynamics (AD) term to represent the dynamics within each clip.
We define the FL term as follows:
\begin{equation}
\mathcal{F}(\mathcal{A}) = \frac{1}{\mathcal{N_{A}}}\sum_{v_i\in \mathcal{A}}{\sum_{v_j\in\mathcal{V}}}d(v_i, v_j), 
\end{equation}
where $\mathcal{N_A}$ indicates the number of the selected facilities. 
In this function, $d(v_i, v_j)$ is defined as (\ref{eq:distance}), which represents the pairwise relation between the candidate facility $v_i$ and the previous selected element $v_j$.
In addition, we formulate the AD term as:
\begin{equation}
\mathcal{U}(\mathcal{A}) = \sum_{v_i\in \mathcal{A}}\exp({-{\phi_i}}),
\end{equation}
where $\phi_i$ is the dynamics score of $v_i$ defined in Section \ref{section:coherence}.
{\flushleft {\bf Optimization for Video Clips Ranking.}}
We combine the FL and AD terms to formulate the submodular problem:
\begin{align}
\underset{\mathcal{A}}{\max} \: \mathcal{L(A)} &
= \underset{\mathcal{A}}{\max} \: \mathcal{F(A)} + \gamma \, \mathcal{U(A)}, \notag\\
& \mbox{s.t.} \; \; \mathcal{A} \subseteq \mathcal{V}, \: \mathcal{N_A} =
N,
\label{eq:sub_combine}
\end{align}
where $\gamma$ is the parameter to balance the contribution of two terms.
The proposed submodular function ensures that the selected facilities share high coherence and maintain rising activity dynamics. 

\begin{figure}\footnotesize
	\centering
	\includegraphics[width=0.95\linewidth]{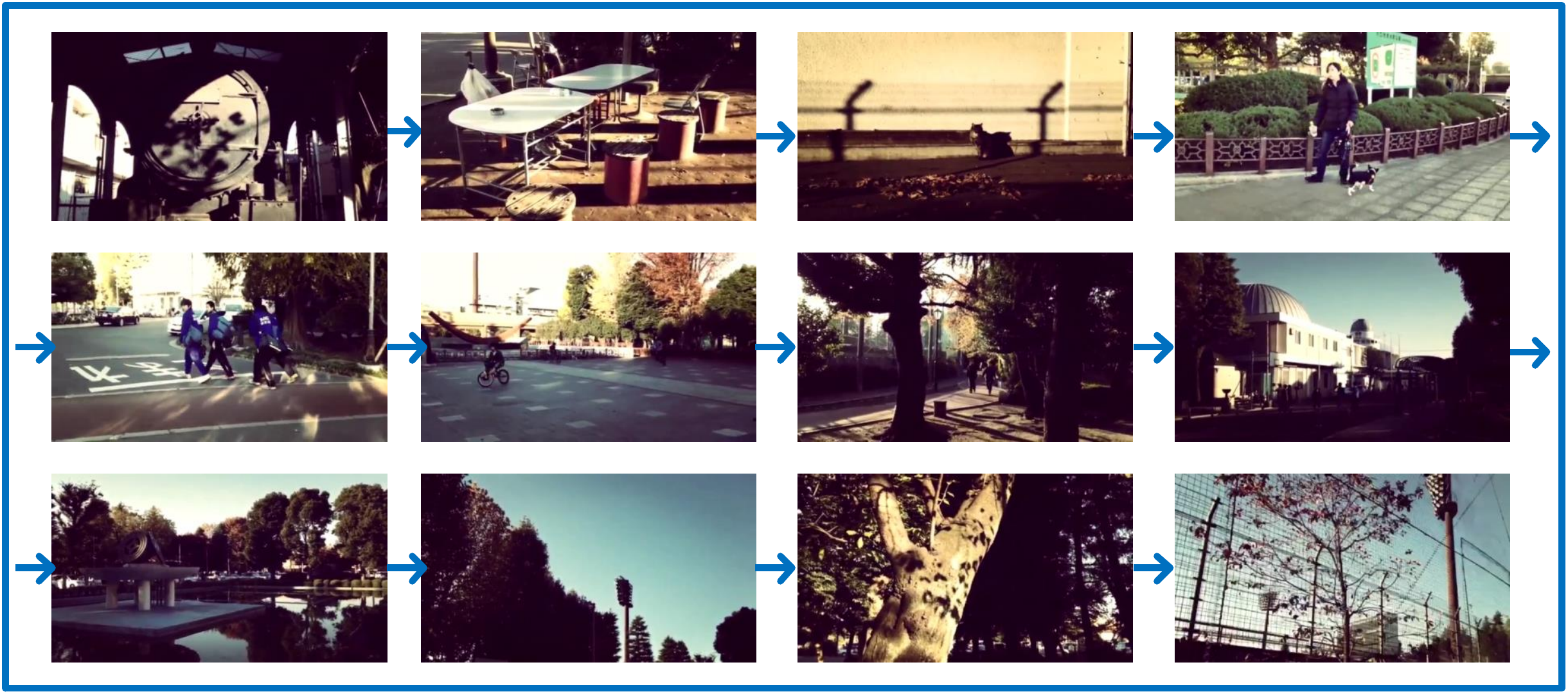}\\
	(a) Video-story generated by our baseline.
	\includegraphics[width=0.95\linewidth]{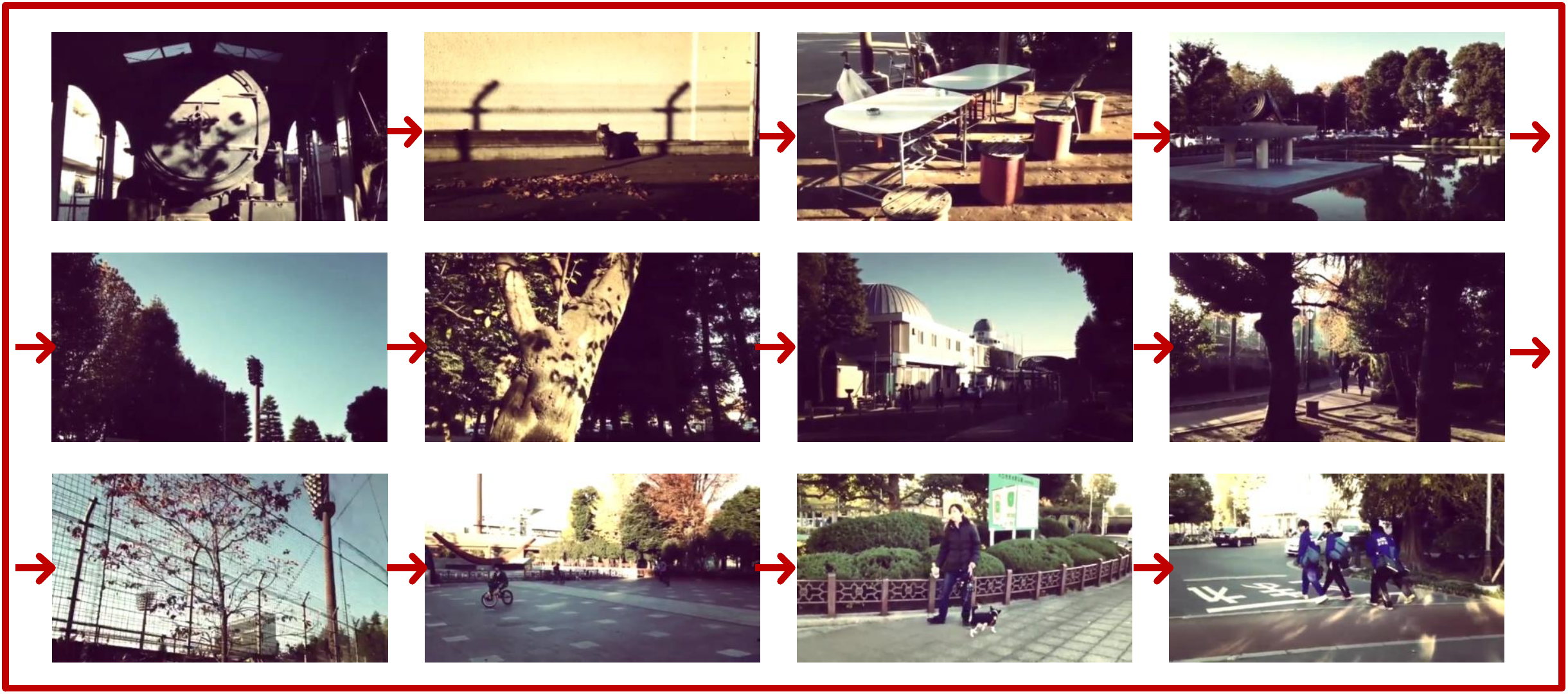}\\
	(b) Video-story generated by the proposed method.
	\includegraphics[width=0.95\linewidth]{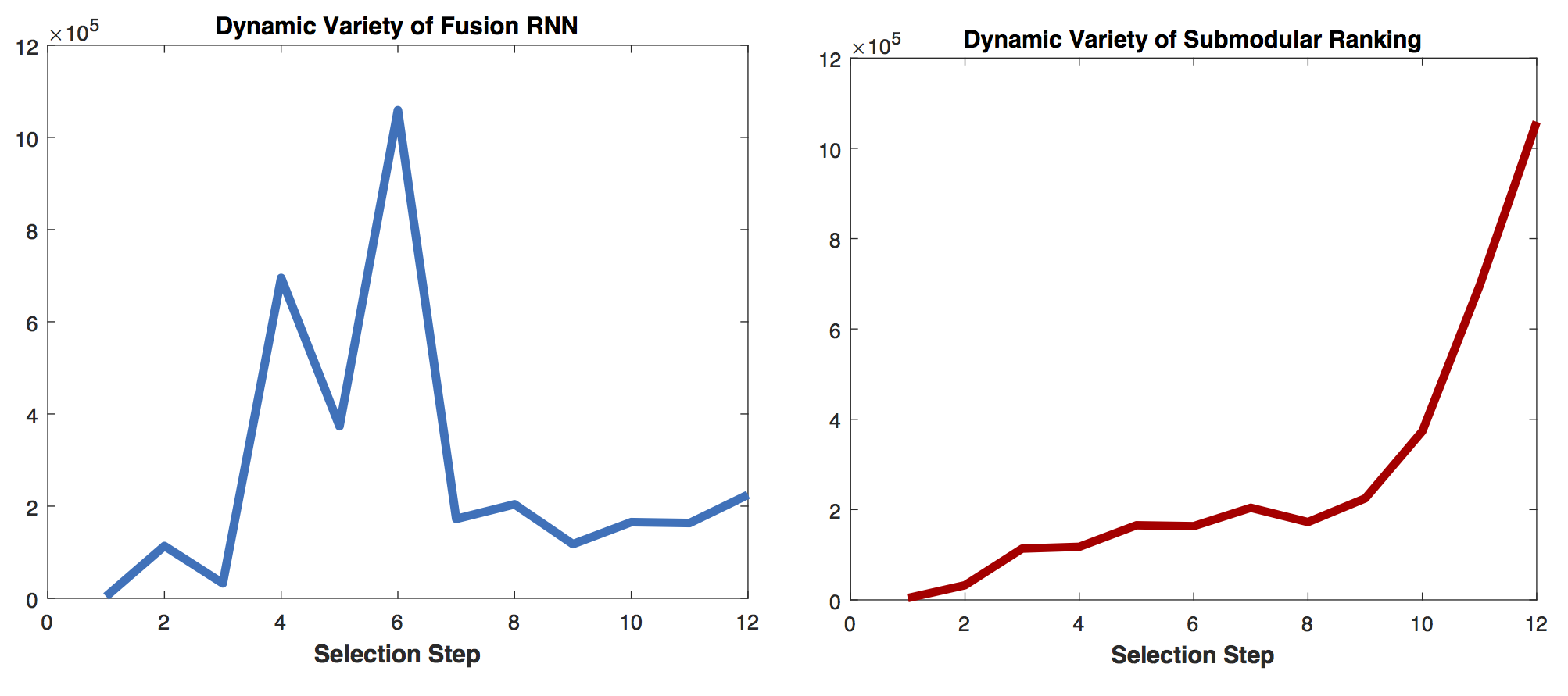}\\
	(c) Dynamics scores of the composed video-story for different methods.
	\caption{(a) and (b) are video-stories generated by the baseline (i.e., two-stream RNN) and proposed method (i.e., two-stream RNN + submodular ranking) where the composition orders are shown by arrows. 
		(c) Dynamics scores of (a) (left) and the proposed method (b) (right). 
		After rearranging the results from (a), our video-story composition result (b) maintains rising dynamics. }
	%
	\label{fig:sub}    
	\vspace{-2mm} 
\end{figure}

As the proposed objective function in \eqref{eq:sub_combine} is the non-negative linear combination of two submodular terms, we solve it using a greedy algorithm similar to \cite{Zhu_CVPR_2014, tsai2016semantic, zhong2016weakly}. 
Since the video-story starts from the exposition with low activities, the facility set $\mathcal{A}$ is first initialized as the node $v_1$ that contains the lowest dynamics score.
Then at the $i$-th iteration, we add the element $a \in \mathcal{V} \setminus \mathcal{A}^{i-1}$ which leads to the maximum energy gain $\mathcal{J}(A^i)$ into $\mathcal{A}$, where the energy gain is defined as: $\mathcal{J}(A^i) = \mathcal{L}(A^i) - \mathcal{L}(A^{i-1})$.
We iteratively select the remaining elements until all the nodes in $\mathcal{V}$ have been selected.
In addition, we use an evaluation form to speed up the optimization process as proposed in \cite{leskovec_sigkdd_2007}. 
The process of submodular ranking is presented in Algorithm \ref{algo:sub}.
Figure \ref{fig:sub} shows the efficiency of our submodular ranking process. By rearranging the baseline results, the video composition contains rising dynamics scores and matches the storyline structure. 
\begin{algorithm}[!t]
	\caption{Optimization for Video Clips Ranking}
	\begin{algorithmic}
		\STATE \textbf{Input}: $G = (\mathcal{V},\mathcal{E}), \mathcal{N}, \gamma$
		\STATE \textbf{Initialization}: $\mathcal{A}^0 \leftarrow \{v_1\}$, $i \leftarrow 1$
		
		\LOOP
		\STATE $a^* = \underset{\{\mathcal{A}^i \subseteq \mathcal{V}\}}{\text{arg max}}$ $\mathcal{J}(\mathcal{A}^i)$, \text{where} $\mathcal{A}^i = \mathcal{A}^{i-1} \cup a$
		
		\IF {$\mathcal{N_A} = \mathcal{N}$}
		\STATE break
		\ENDIF
		
		\STATE $\mathcal{A}^i \leftarrow \mathcal{A}^{i-1} \cup a^*$, 
		\STATE $i=i+1$
		
		\ENDLOOP
		
		\STATE \textbf{Output}: $\mathcal{A} \leftarrow \mathcal{A}^i$
	\end{algorithmic}
	\label{algo:sub}
\end{algorithm}
\vspace{-2mm} 
\section{Experimental Results}	
We evaluate the video-story composition results in this section. We first introduce the dataset and experimental details in Section \ref{sec:experimental_details} and then analyze the quantitative and qualitative results in Section \ref{sec:experimental_result}.	
%
%
%
\begin{figure*}\footnotesize
	\centering
	\includegraphics[width=0.95\linewidth]{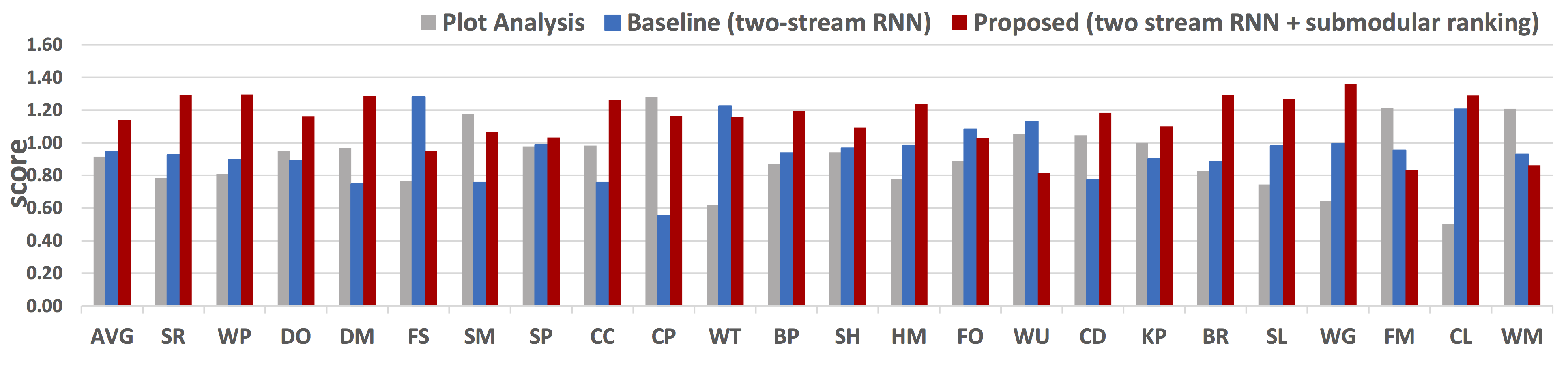}\\
	\caption{Pairwise preference scores. The scores are generated in a similar way to \cite{choi_CVPR_2016_video}. 
		The proposed method (i.e., two-stream RNN + submodular ranking) receives higher average pairwise score (i.e., 1.14), compared with our baseline (i.e., two-stream RNN) and the PA \cite{choi_CVPR_2016_video} method with the average pairwise scores of 0.94 and 0.91, respectively.
		The abbreviations indicate the contents of each video set: SR (surfing + river), WP (walk + park), DO (drive + ocean), DM (drive + morning), FS (family + swim), SM (shopping + mall), SP (sightseeing + park),
		CC (chatting + cafe), CP (couple + park), WT (walk + trees),
		SP (skateboarding + park), SH (sightseeing + hill), HM (hiking + mountain),
		FO (friend + ocean), WU (walk + urban), CD (cat + dog), KP (kid + park),
		BR (skateboard + road), SL (sightseeing + lake), WG (walk + garden),
		FM (family + market), CL(car + lake), WM (walk + museum).
	}
	\label{fig:bar}   
	\vspace{-2mm} 
\end{figure*}
\subsection{Experimental Details}
\label{sec:experimental_details}
{\flushleft {\bf Datasets.}} 
We evaluate the proposed method on the video composition dataset \cite{choi_CVPR_2016_video}
which consists of 23 video sets collected from YouTube. Each video set contains 8-12 video clips which last for 2-3 seconds, and the whole dataset has 236 video clips. The dataset contains rich activity contents (e.g., sightseeing, skateboarding, walking, surfing, shopping, driving, and swimming) in various scenes (e.g., river, park, ocean, streets, mall, landmarks, museum, marketplace, garden, and beach). 

We train the models on the SumMe \cite{gygli_ECCV_2014,gygli_CVPR_2015} and TVSum \cite{caba2015activitynet} datasets that consist of 25 videos (with the average length of 160 seconds) and 50 videos (with the average length of 252 seconds) respectively. The training sets cover the activity contents of holidays, events and sports. 
The videos in the datasets have good consistency and are suitable for learning video coherence as formulated in the proposed RNN.

{\flushleft {\bf Experimental Settings.}} 
In the process of learning video coherence, considering the content varieties and lack of ground truths, we train the RNNs in an unsupervised manner. 
We use a fixed number (i.e., $T = 10$ in this work) of temporally continuous clips as a training sequence. Each item in the sequence is a video clip with 16 frames. 
The input of the semantic stream is a 4096-dimensional $fc7$ feature from the C3D model. 
To describe motion contents, we set the bin number of the HOOF feature as 10 and set the pyramid level as $\{3 \times 3\}$, resulting in the input size as 100. The hidden recurrent layer size is set to 100. Both the  C3D and SPP-HOOF features are directly fed into the models.

We set $\sigma_h$ and $\sigma_y$ as the activation function of the rectified linear unit \cite{Krizhevsky_NIPS_2012} and set the momentum of the gradient ascent as 0.9. 
We start the training process with the learning rate as 0.05, and gradually reduce it till the likelihood no longer increases with the weight decay $\lambda = 10^{-7}$. 
%
During the test phase, we initialize the RNN with the first video clip that has the lowest dynamics score as mentioned in Section \ref{section:coherence} and set $\gamma =  0.3$ in \eqref{eq:sub_combine}.
%
\subsection{Evaluation Results}
\label{sec:experimental_result}
\begin{figure}\footnotesize
	\centering
	\begin{tabular}{c@{\hspace{0.1mm}}c@{\hspace{0.1mm}}c@{\hspace{0.1mm}}c@{\hspace{0.1mm}}c}
		\includegraphics[width=0.5\linewidth]{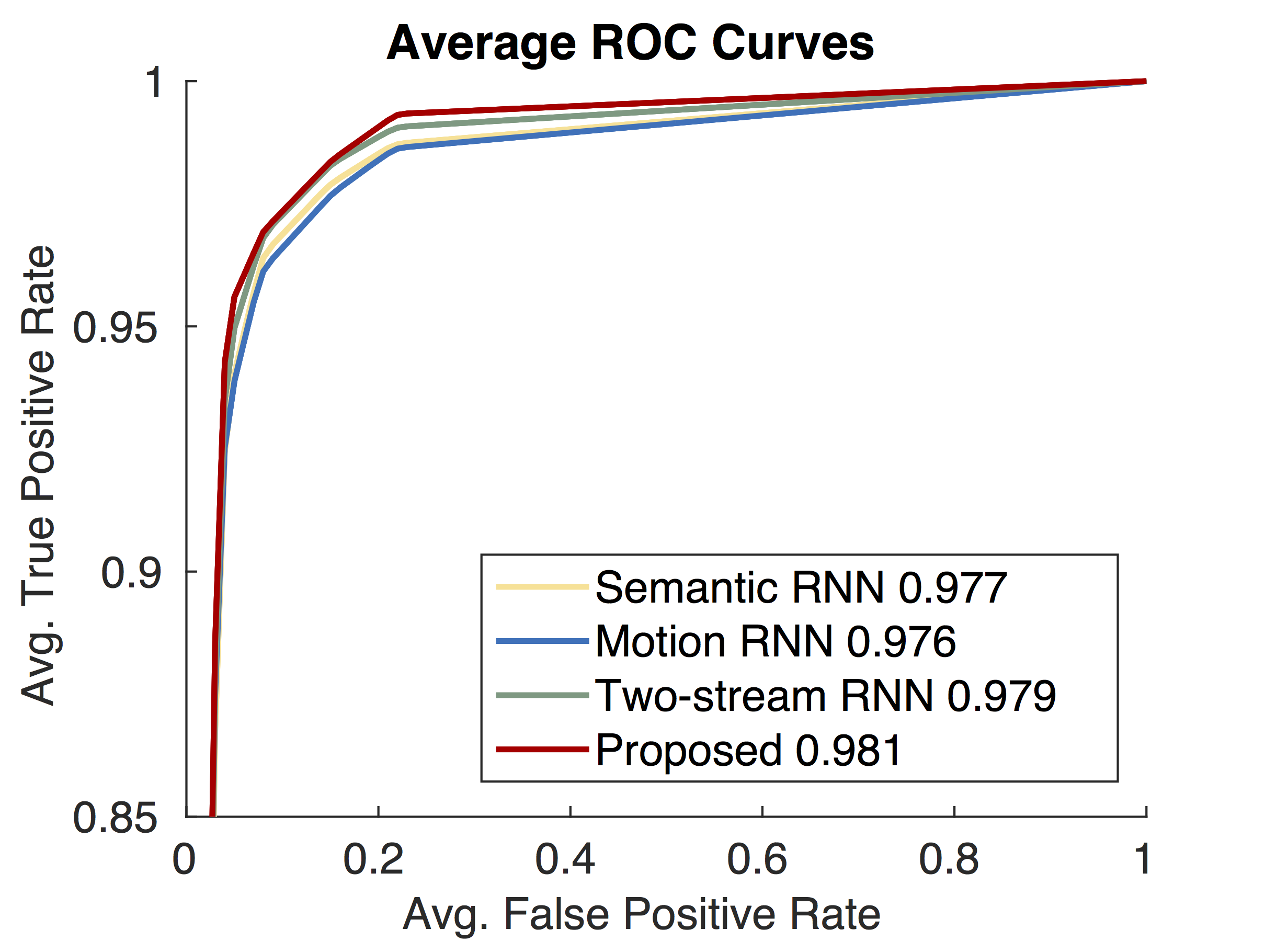}&
		\includegraphics[width=0.48\linewidth]{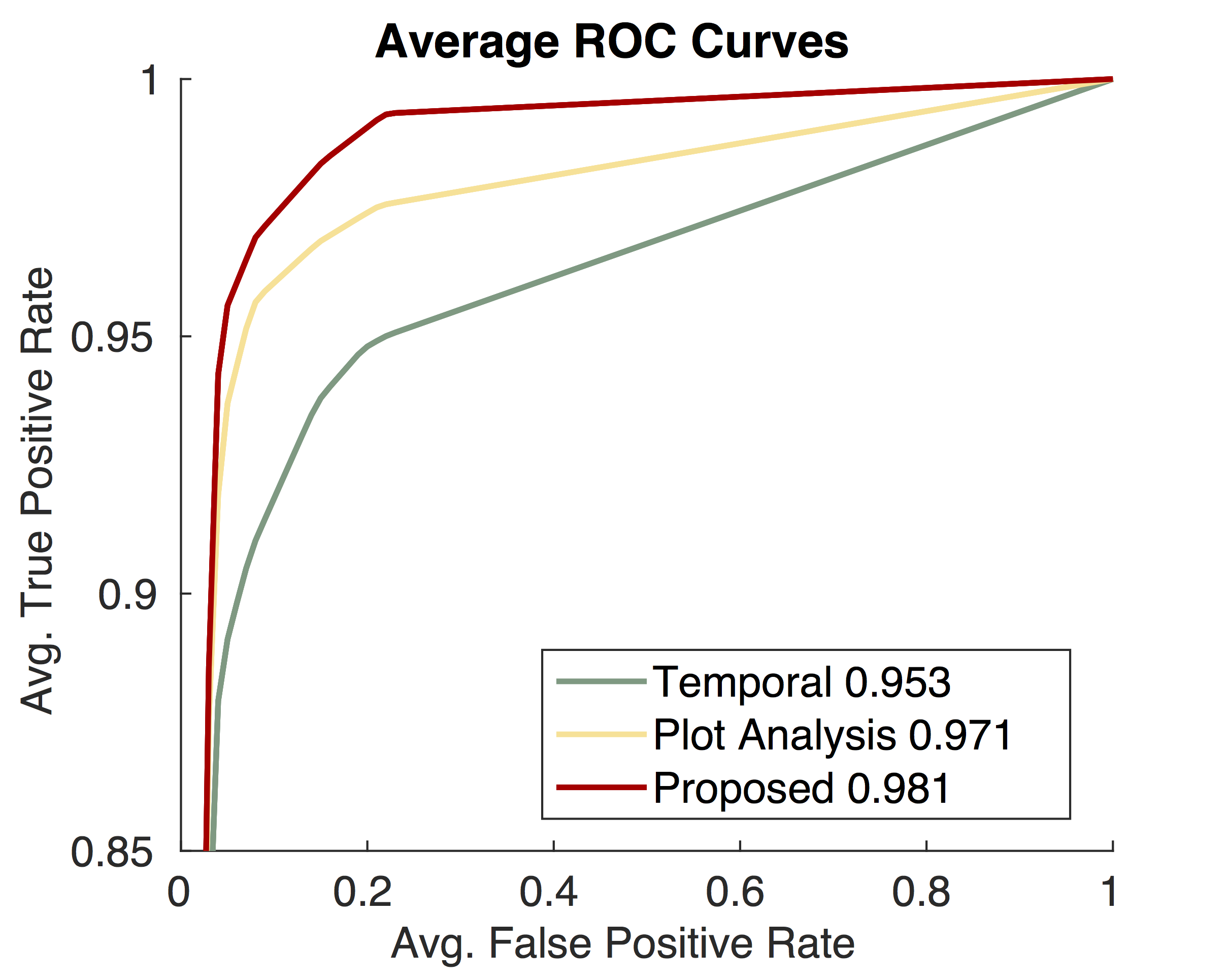}\\
		(a) & (b) \\
	\end{tabular}
	\caption{Component-wise comparison results. The corresponding AUC scores are also provided in the legend. 
		(a) average ROC curves for our methods with different design options.
		(b) average ROC curves for the proposed method and the comparisons. Our method achieves higher ROC curves and AUC scores compared to several baseline methods and the state-of-the-art algorithm.
	}
	\label{fig:curve}
	\vspace{-2mm} 
\end{figure}	
We evaluate the video-story composition results in global and local aspects, i.e., overall video-story quality and component coherency.
We first analyze the efforts of the two streams (i.e., the semantic RNN and the motion RNN) in our framework. We then compare the proposed method (i.e, two-stream RNN + submodular ranking) and our baseline (i.e., two-stream RNN) against the state-of-the-art method, i.e., Plot Analysis (PA) \cite{choi_CVPR_2016_video}. More experimental results can be found in the supplementary material, and the MATLAB codes will be available on \url{https://github.com/GYZHikari/VideoStory_RNN}.

\begin{figure*}\footnotesize
	\centering
	\includegraphics[width=0.95\linewidth]{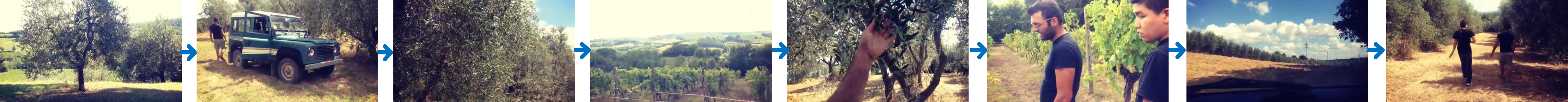}\\
	(a) Video-story on WG generated by PA \cite{choi_CVPR_2016_video}.
	\includegraphics[width=0.95\linewidth]{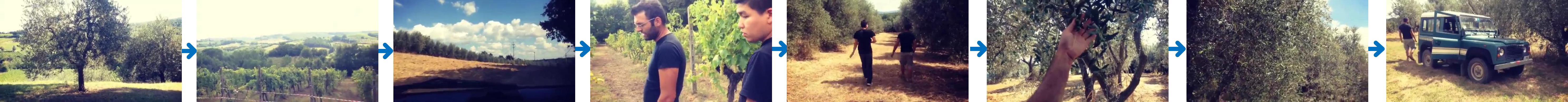}\\
	(b) Video-story on WG generated by our baseline (i.e., two-stream RNN).
	\includegraphics[width=0.95\linewidth]{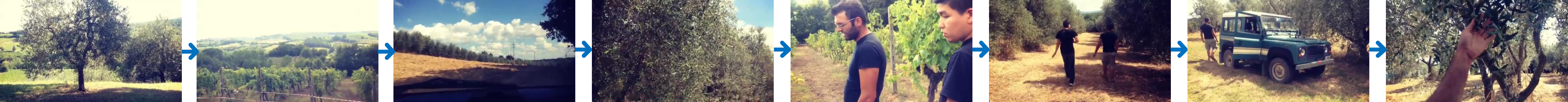}\\
	(c) Video-story on WG generated by the proposed method (i.e., two-stream RNN + submodular ranking).
	\caption{Video-story results on the WG video set. The video clips in this video set contain various scenes and human actions. 
		Our baseline and the proposed method order the similar scenes and activities together while the results generated by PA \cite{choi_CVPR_2016_video} show less consistency. 
	}
	\label{fig:WG}
	\vspace{-2mm} 
\end{figure*}

\begin{figure*}\footnotesize
	\centering
	\includegraphics[width=0.95\linewidth]{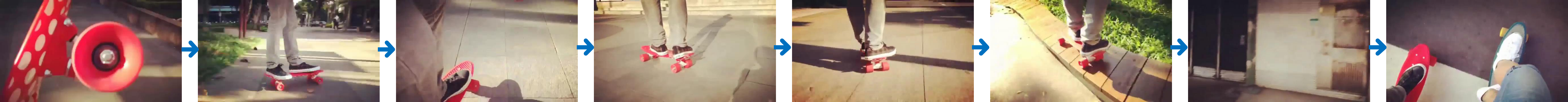}\\
	(a) Video-story on BR generated by PA \cite{choi_CVPR_2016_video}.
	\includegraphics[width=0.95\linewidth]{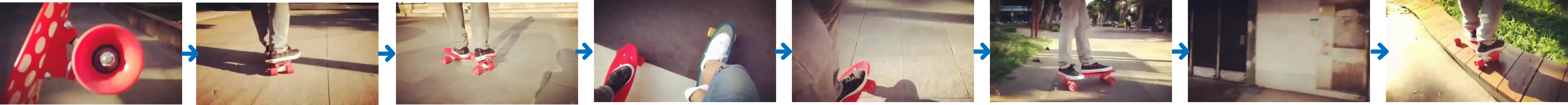}\\
	(b) Video-story on BR generated by our baseline (i.e., two-stream RNN).
	\includegraphics[width=0.95\linewidth]{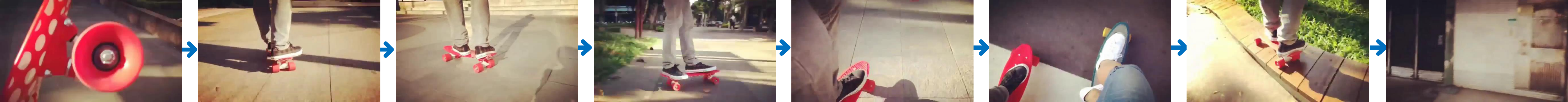}\\
	(c) Video-story on BR generated by the proposed method (i.e., two-stream RNN + submodular ranking).
	\caption{Video-story results on the BR video set. The video clips in this video set contain similar activities and various dynamics. The proposed baseline method produces smooth transitions, while our submodular ranking process further improves the baseline results.
	}
	\label{fig:BR}
	\vspace{-2mm}  
\end{figure*}

{\flushleft {\bf Overall Video-Story Quality.}}
Since evaluating the quality of the video-story composition is complex and subjective, we conduct a user study on the Amazon Mechanical Turk following the settings used in \cite{choi_CVPR_2016_video}.
We show each subject to choose the one with the better video-story from a pair of composed results.
Each pair consists of video-stories composed by two different methods while containing the same contents.
All the video-story results from the dataset are shown in random orders.
Our evaluation involves 134 subjects, resulting in a total of 3,105 pairwise results.
After obtaining all the pairwise results, we use the Bradley-Terry (B-T) model \cite{bradley_1952_rank, Lai_CVPR_2016} to obtain the global ranking scores.
The B-T scores of the  proposed algorithm, our baseline and the Plot Analysis  \cite{choi_CVPR_2016_video} method are 1.22, 0.88 and 0.85, which demonstrates 
the effectiveness of the proposed model.

Similar to \cite{choi_CVPR_2016_video}, Figure \ref{fig:bar} shows the results of the pairwise preference test of our method and the comparisons. 
We find that the proposed method performs better when the given clips contain various scenes and activities, e.g., BR and WG video sets.
%
%
In our method, the learned video coherence can help to handle such challenges and generate results with smooth and consistent view transitions.
Figure \ref{fig:WG} shows another example where the WG video set contains walking and garden such that the appearances (e.g., color distribution) in the relevant clips are similar.
As a result, the PA method does not perform well due to ambiguous coherence. 
In contrast, the proposed RNN models the relations of scenes and activities, and thus generate video-stories with more coherent contents.
In addition, we demonstrate the proposed submodular ranking process can further improve results in a dynamic scene environment.
In Figure \ref{fig:BR}, the BR video set consists of a series of skateboarding actions. 
The scenes between video clips are similar (e.g., the road) and the contents change dynamically.
The results show that our submodular ranking process incorporates both the dynamics and coherence to generate a smoother video-story.

{\flushleft {\bf Component Composition Quality.}} 
\begin{figure}\footnotesize
	\centering
	\includegraphics[width=0.95\linewidth]{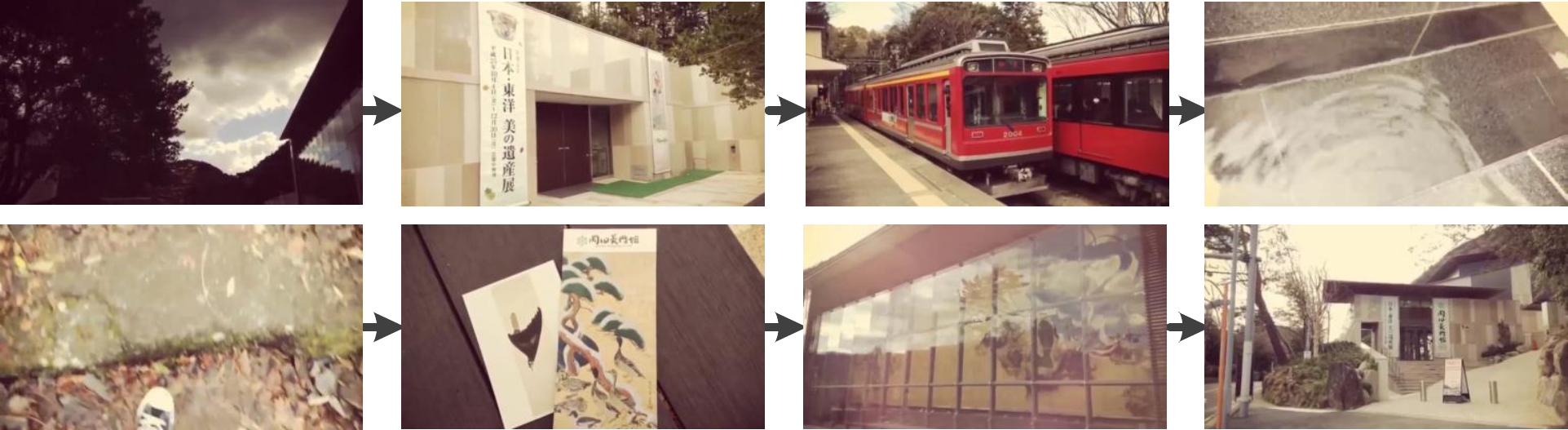}
	 \vspace{-2mm}
	\caption{Failure cases by the proposed method.}
	\label{fig:fail}
	\vspace{-3mm} 
\end{figure}
In this task, we evaluate the coherence quality between two adjacent components in the composed video-story. We evaluate the results using the component-wise ground truths provided from \cite{choi_CVPR_2016_video}.
%
For each video set, we obtain the average ROC curves generated by the proposed algorithm and evaluated methods. 
We first evaluate different components of our framework, i.e., semantic RNN, motion RNN, two-stream RNN (our baseline), and the submodular ranking process. 
Figure \ref{fig:curve}(a) shows that our baseline, i.e., two-stream RNN, incorporates both semantic and motion information, and generate better component-wise results. 
In addition, our submodular ranking process further improves the performance, in which the dynamics are considered to better match the video-story structure.
Then we compare the proposed method with the state-of-the-art method, i.e., PA \cite{choi_CVPR_2016_video} and the temporal results in Figure \ref{fig:curve}(b).
Our method achieves higher ROC curve and AUC scores, which shows the effectiveness of our learned coherence and submodular ranking process.
{\flushleft {\bf Failure Cases.}}
As shown in Figure \ref{fig:fail}, in the cases with uneventful or irrelevant scenes and activities 
(e.g., walk and museum),  our method shows less effectiveness since the contents of clips do not affect the whole story.

\section{Concluding Remarks}
In this paper, we focus on the video-story composition task via a learning based approach. 
Since the video contents may change significantly through the time, we exploit the coherence between video clips to predict connections for compositing video-stories.
In the proposed framework, we train the two-stream RNN in terms of spatial-temporal semantics and motion dynamics. 
The probabilities generated by the two-stream RNN are fused as the coherence scores of video clips to generate smooth and relevant compositions.
To further match the video-story structure, we formulate a submodular ranking problem to rearrange the video-story composition.
Experimental results on the video-story dataset show that the proposed algorithm performs favorably against the state-of-the-art approach.
{\flushleft {\bf Acknowledgements:}} This work is supported in part by NSFC (No. 61572099 and 61522203), NSF CAREER (No. 1149783), 973 Program (No. 2014CB347600), NSF of Jiangsu Province (No. BK20140058), the National Key R\&D Program of China (No. 2016YFB1001001), and gifts from Adobe and Nvidia.
\clearpage
{\small
	\bibliographystyle{ieee}
	\bibliography{video_story}
}

\end{document}